%% file: neurips_2025.tex
\documentclass{article}

      \usepackage[preprint]{neurips_2025}

\usepackage[utf8]{inputenc} 
\usepackage[T1]{fontenc}    
\usepackage{hyperref}       
\usepackage{url}            
\usepackage{booktabs}       
\usepackage{amsfonts}       
\usepackage{nicefrac}       
\usepackage{microtype}      
\usepackage{graphicx}

\usepackage[table]{xcolor} 
\usepackage{booktabs}      
\usepackage{array}         
\usepackage{enumitem}      

\newcommand{\datasetname}[1]{MONITRS#1}
\usepackage{multirow}
\usepackage{multicol}

\usepackage{tcolorbox}
\newtcolorbox{promptbox}[1][]{
    colback=blue!5,
    colframe=blue!60!black,
    boxrule=1.5pt,
    left=8pt,
    right=8pt,
    top=8pt,
    bottom=8pt,
    fonttitle=\bfseries,
    title=#1,
    #1
}

\usepackage{subcaption} 
\title{MONITRS: Multimodal Observations of Natural Incidents Through Remote Sensing}

\author{
Shreelekha Revankar$^{1}${\thanks{Corresponding Email: revankar@cs.cornell.edu}}, Utkarsh Mall$^{2}$, Cheng Perng Phoo$^{1}$, Kavita Bala$^{1}$, Bharath Hariharan$^{1}$\\
\\
$^{1}$Cornell University
$^{2}$Columbia University
}

\begin{document}

\maketitle

\begin{abstract}
Natural disasters cause devastating damage to communities and infrastructure every year. Effective disaster response is hampered by the difficulty of accessing affected areas during and after events. Remote sensing has allowed us to monitor natural disasters in a remote way. More recently there have been advances in computer vision and deep learning that help automate satellite imagery analysis, However, they remain limited by their narrow focus on specific disaster types, reliance on manual expert interpretation, and lack of datasets with sufficient temporal granularity or natural language annotations for tracking disaster progression. We present MONITRS, a novel multimodal dataset of more than 10,000 FEMA disaster events with temporal satellite imagery and natural language annotations from news articles, accompanied by geotagged locations, and question-answer pairs. We demonstrate that fine-tuning existing MLLMs on our dataset yields significant performance improvements for disaster monitoring tasks, establishing a new benchmark for machine learning-assisted disaster response systems.
\end{abstract}
\input{sections/introduction}
\input{sections/related_works}
\input{sections/methodology}
\input{sections/experiments}

\input{sections/conclusion}

{
\small

\bibliography{biblio}
\bibliographystyle{plain}
}



\input{sections/appendix}

\end{document}

%% file: sections/introduction.tex
\section{Introduction}
\input{figures/intro_fig}
Natural disasters cause significant damage to infrastructure, homes, and communities, resulting in loss of life and billions of dollars in economic costs annually. Effective disaster response depends on understanding what events are occurring, where they are taking place, and how they progress over time~\cite{fema_dis}. However, affected regions are often inaccessible or dangerous to access during and after disasters. 

A promising solution is automatic analysis of satellite imagery, enabling non-invasive coverage of disaster zones~\cite{chen2024application}. However, natural disasters pose unique challenges for such analysis: they are characterized by rapid change in a short period of time, and understanding this rapid temporal evolution is critical for disaster management. Unfortunately, much of the recent literature on recognizing concepts in satellite imagery focuses on static concepts like land-use and is not equipped to analyze rapid change events like natural disasters. Approaches that do detect change often do not allow for semantic interpretation~\cite{yang2024made} or do not provide fine-grained temporal understanding~\cite{dong2024changeclip, hoxha2022change, irvin2024teochat}. The few approaches that have been proposed specifically for natural disasters either focus on specific disaster types with specialized models~\cite{tanim2022flood,braik2024automated} or require substantial manual interpretation by domain experts~\cite{galetto2024use}. 

A key challenge in building recognition models for disaster understanding is the lack of annotated datasets. However, building such a dataset is difficult: natural disasters are by definition rare, and straightforward sampling of remote-sensing imagery is unlikely to chance upon these events. Even if we were to get remote sensing imagery from natural calamities, they are not annotated with the kinds of concepts we may want recognized. For instance, many of the available annotations for satellite imagery revolve around land-use, which is why existing approaches can recognize when buildings are built, but not where wildfire scarring has occurred. This lack of annotations cannot be resolved easily through manual annotations because remote sensing imagery is an unfamiliar domain for most lay annotators.

In this paper, we address this data challenge by presenting \datasetname{} (\textbf{M}ultimodal \textbf{O}bservations of \textbf{N}atural \textbf{I}ncidents \textbf{T}hrough \textbf{R}emote \textbf{S}ensing) --- a first-of-its-kind dataset of remote-sensing imagery of natural disasters annotated with natural language descriptions. 
Our key insight is to pair public records of natural disasters in the US maintained by the Federal Emergency Management Agency (FEMA) with \emph{news articles} covering these events and containing detailed natural language descriptions. We propose a novel data curation pipeline that combines these sources to produce a unified resource for disaster monitoring research and application development.
 
 \datasetname{}~consists of approximately 10,000 disaster events documented by FEMA, paired with: 

\begin{itemize}[itemsep=0pt, parsep=0pt]
    \item Temporal sequences of geolocated satellite imagery capturing each event's progression,
    \item Natural language annotations derived from news articles describing the events,
    \item Precise geotagged locations marking areas of interest within each event, and finally
    \item Question-answer pairs designed to train and evaluate multimodal language models
\end{itemize}
Unlike existing disaster monitoring datasets that focus on single disaster types or limited temporal windows, \datasetname{}~captures the complete lifecycle of diverse disaster events, from initial impact through recovery phases.

Using our dataset, we demonstrate that existing remote-sensing multimodal LLMs (mLLMs) are indeed unable to understand the progression of natural disasters. We find that existing models are particularly bad at temporal grounding and event classification for natural disasters. To address these limitations, we fine-tune existing MLLMs on our dataset and demonstrate improved performance in the domain of disaster response. 


Our work addresses a significant gap in disaster monitoring resources and lays the groundwork for more effective, machine learning-assisted disaster response systems that combine the geographic comprehensiveness of satellite imagery with the accessibility of natural language interfaces.


%% file: figures/intro_fig.tex
\begin{figure}[ht]
    \centering
    \includegraphics[width=0.9\linewidth]{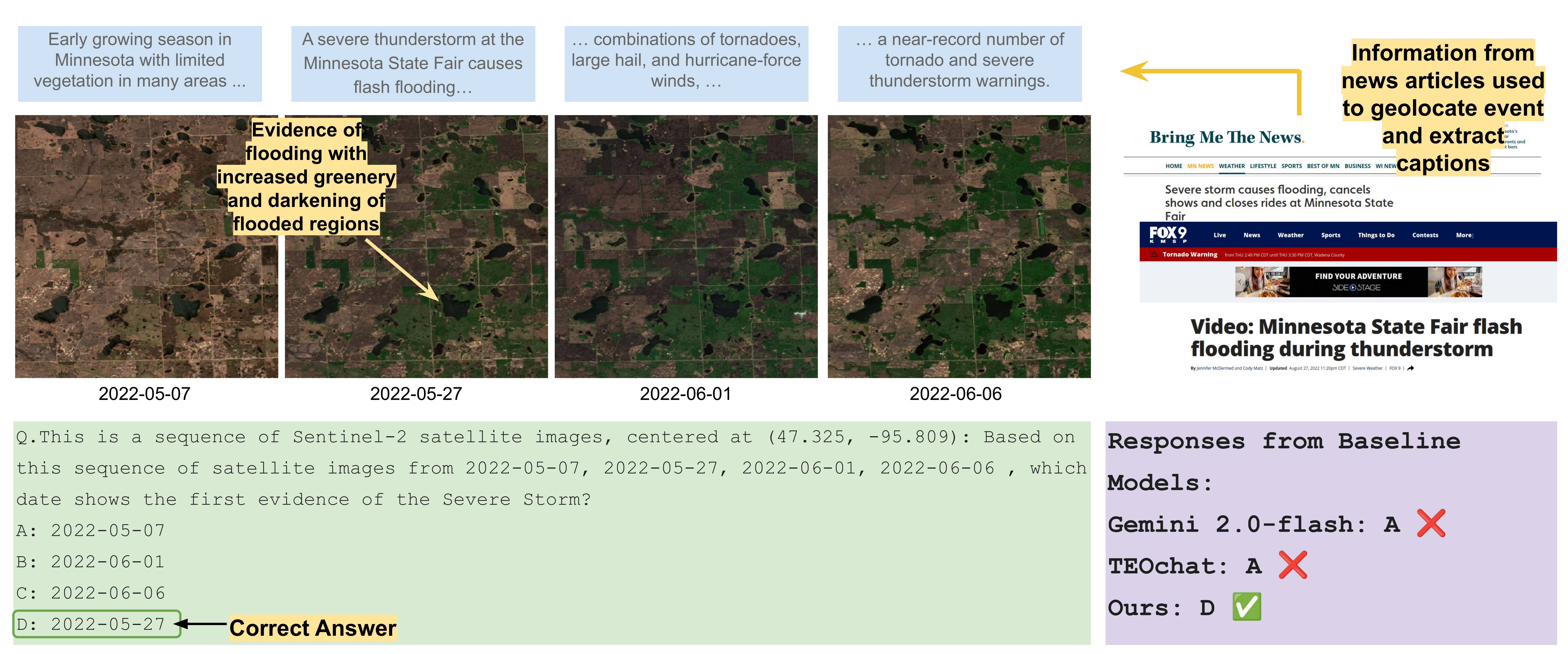}
    \caption{Using news articles, we extract exact locations of disaster events and corresponding captions for event timelines. Our MONITRS dataset enables precise disaster monitoring, as shown in this Minnesota severe storm sequence. The May 27th image shows evidence of flooding with increased vegetation and darker water-saturated regions. Models finetuned with MONITRS correctly identify the temporal onset of the storm while baseline models fail to detect the initial evidence.}
    \label{fig:enter-label}
\end{figure}

%% file: sections/related_works.tex
\section{Related Works}

\subsection{Event Monitoring using Earth Observation Data}
Many ML methods have been used to model temporal sequences of earth observation data. Particularly in disaster monitoring, automated methods for change detection can help in planning disaster relief, assessing damage extent, and monitoring recovery. These approaches typically analyze pairs or sequences of images capturing the same location over time to identify changes that indicate disasters~\cite{sachdeva2023change,yang2024made,noman2024remote}.

Disaster monitoring presents unique challenges compared to general change detection tasks, as changes can be sudden and dramatic and require models that can distinguish between normal changes (for example, seasonal changes) and disaster-induced ones~\cite{park2024national,mall-23,manas2021seasonal}. Prior works have explored various approaches for disaster-specific applications, including building damage assessment~\cite{braik2024automated}, flood extent mapping~\cite{tanim2022flood}, wildfire tracking~\cite{thangavel2023autonomous}, and post-disaster recovery monitoring~\cite{sheykhmousa2019post}. However, most existing approaches are designed for specific disaster types or short temporal windows. This limits the types of disasters that any one system can monitor~\cite{said2019natural}.

While change detection techniques have made significant progress in identifying visual differences between temporal imagery, they typically lack natural language understanding capabilities~\cite{mall-23,manas-21}. Some specialized models can identify and distinguish certain events, but they can only process limited time sequences, making them insufficient for comprehensive disaster monitoring that requires tracking changes over extended periods~\cite{dong2024changeclip,irvin2024teochat,hoxha2022change}.

\subsection{Vision-Language Models for Earth Observation Data}
Efforts to develop VLMs for EO data have been rapidly increasing. These methods commonly use different single-image EO datasets and convert them to instruction-following tasks, then fine-tune a LLaVA-like model on the dataset~\cite{kuckreja2024geochat,irvin2024teochat}.

Recent works have introduced novel image-caption datasets for training remote sensing foundation models, pairing aerial and satellite imagery with captions generated using landmarks or utilizing public web images with the text filtered for the remote sensing domain~\cite{revankar2024scale,mall2023remote,liu2024remoteclip}. These approaches have demonstrated state-of-the-art generalization performance in zero-shot retrieval. 

Most existing VLMs for Earth Observation are designed to handle single image inputs, limiting their use for many real-world tasks that require temporal reasoning, particularly for phenomena like natural disasters that evolve over time~\cite{li2024vision}.

Several recent works have developed VLMs that can engage in conversation about videos, demonstrating the potential for temporal reasoning in multimodal models~\cite{lin2023video,zhang2023video}. Approaches such as TEOChat~\cite{irvin2024teochat} have shown that video-language models can be adapted to handle temporal sequences of earth observation data, performing a wide variety of spatial and temporal reasoning tasks. However, these models are constrained by the lack of temporal granularity in existing training datasets for remote sensing events. This limitation prevents tracking the full progression of natural disasters.

\subsection{Multimodal Datasets for Remote Sensing Events}

Existing multimodal datasets for remote sensing typically focus on a limited set of tasks or specific disaster types~\cite{liu2022remote,zhu2024semantic}. Various change detection datasets focused on building change~\cite{gupta2019xbd,braik2024automated}, land cover changes, or land use changes~\cite{zhu2024semantic}. While several works have designed self-supervised approaches to leverage temporal sequences of earth observation data~\cite{yang2024made,manas2021seasonal,mall-23}, few have developed comprehensive datasets that combine satellite imagery, geospatial information, and textual annotations derived from real-world sources like news articles.

The lack of large-scale, diverse datasets that include multiple disaster types, temporal scales, and annotations, presents a significant bottleneck for developing general-purpose models for disaster monitoring and response. Our work addresses this gap by creating a comprehensive dataset covering approximately 10,000 disaster events from FEMA, incorporating geolocated satellite imagery throughout the duration of events, natural language annotations from news articles, geotagged locations relevant to the events, and question-answer pairs for training multimodal language models.

%% file: sections/methodology.tex
\section{\datasetname{}}
Effective monitoring of natural disasters requires us to understand certain details about the disaster, such as where it is occurring, when it began, and how it affects the infrastructure and communities in its path. We aim to automate this process via satellite imagery so that we can perform effective monitoring over large areas in a non-invasive, less labor intensive way.

Recent works have demonstrated that large multimodal language models can act as powerful tools for understanding events~\cite{irvin2024teochat,lin2023video}. However, current datasets do not capture the necessary details to train such a model to act as a sufficient tool for the task at hand. We create a novel natural disaster dataset that captures the required information.
\input{figures/geolocating_articles3}
\input{figures/caption_construction}

\subsection{\datasetname{} Construction}
The first challenge we need to address is the relative rarity of natural disasters. As such, simply sampling remote sensing imagery is unlikely to yield enough samples for these events.
Instead, we begin with FEMA's Disaster Declarations Areas~\cite{fema_set}, which includes a list of all federally declared disasters. This helps us define the types of disasters we include in our scope. 
Since we want to acquire the relevant satellite imagery that tracks each event, we only keep events that have enough information to spatio-temporally localize the event, namely, county, state, event name, and start and end dates. Events that do not have this information are discarded.

While FEMA keeps some information of the disasters, they do not keep detailed descriptions of their extent. For example, while the records contain the county where the disaster occurred, the true locations of the disaster and its effects can be far from the exact centers of these counties. 
This poses a challenge in acquiring the right remote-sensing imagery that captures the full extent of the event. 
In addition, the FEMA database does not include any annotations or descriptions of the evolution of the event, which would be needed to train capable remote-sensing multimodal LLMs. 





\textbf{News articles for events:} We find that a better way to locate the full extent of these events is to leverage news articles written about the disaster. 
These articles provide detailed descriptions that capture which specific regions were affected, when and how. This not only allows us to geolocate the event correctly, but also provides us with natural language descriptions that describe the evolution of the event in detail.

To find relevant news articles, we construct search queries using our filtered list of FEMA events. The queries are comprised of the event name, county, state, and start date. For each event, we collect news articles or reports. To reduce the chance of accidentally including irrelevant information, we select the first five results returned by the search query, using the Google Search API~\cite{GoogleCustomSearchAPI}.

From these articles, we first ascertain the exact location and geographical extent of the natural disaster being reported on. We begin by parsing through the articles using LLMs, specifically the freely available Gemini 2.0-flash model. We ask the model to retrieve all of the proper nouns of locations mentioned in the articles. For example this includes specific highways, or town names. We create a union of all the locations mentioned across the articles and retrieve their geocoded location (latitudinal and longitudinal position) using the Geocoding API~\cite{geocodemapsco}. 
This gives us a more complete representation of the extent of the event.

\textbf{Acquiring satellite images:} With these locations at hand, we select the square patch (of fixed size) that includes the maximum number of proper noun locations mentioned across all articles. 
This square patch forms the basis for acquiring satellite imagery. 
As a source of satellite images, we use RGB bands of Sentinel-2 imagery, which is publicly available~\cite{sentinel}. Sentinel-2 imagery has a ground sampling distance of 10m per pixel and a re-visit rate of 5 days on average. The size of the square patch is $5.12 \times 5.12 km^2$, which corresponds to a 512x512 pixel image.
With this region we download all available satellite images for the duration of the natural disaster as reported by FEMA, including a 10 day buffer before and after the event to ensure we capture its entirety. 

\textbf{Acquiring natural language descriptions:} 
The final step is to produce natural language descriptions of the event. We wish to produce descriptions for the temporal evolution of the event. 
To this end, we make note of all of the dates that comprise the natural disaster event.
We then prompt Gemini with these dates and with the text of all the news articles for the event (which includes dates as well), and ask it to describe what visible events have occurred by each date. 
This is done using the article content and dates alone.

Ultimately, through this process, for a set of natural disaster events we have, (a) the approximated locations of the events, (b) satellite imagery that covers the event, (c) a list of geolocated proper nouns that are affected or associated with the event, (d) detailed descriptions of the event through time captured using (e) news articles reporting on the event.
The five components make up \datasetname, and can be used to support several downstream tasks.

Next, we use this dataset to create a VQA datasets to benchmark and finetune large multimodal language models for answering questions about events from satellite imagery.

\input{figures/dataset_stats}

\subsection{Dataset Statistics}
Our dataset contains 9,996 disaster incidents collected from FEMA records. We visualize statistics about the dataset in Figure~\ref{fig:dataset_stats}. Hurricanes and severe storms constitute the majority of events, with strong seasonal patterns peaking in September. Geographic distribution centers primarily in coastal and hurricane-prone regions, with the states of Louisiana, Texas, and Florida experiencing the highest incident counts. On average there are 4.13 images per event, representing on average 18.14 days.






\section{\datasetname-{QA}}


With \datasetname{}, we have sufficient information to construct a visual question-answering dataset for natural disasters. We utilize two formats of question-answer datasets for different purposes. The first being multiple-choice QA datasets, so that correct answers can be confirmed easily for quantitative results. The second being open-ended QA datasets, which allows for more detailed and descriptive responses.

We develop these datasets using two approaches. The first is templated question and answers, where we standardize questions with slots for event-specific information. Using a template allows us to evaluate model performance for specific kinds of reasoning. The second is generated question and answers, where we employ large language models to create diverse, event specific questions with linguistic variety.

\textbf{Templated questions:} The types of reasoning covered in our templated questions include \textit{event classification}, \textit{temporal grounding}, and \textit{location grounding}:

    


 
\textit{Event Classification} questions ask the model to categorize the event.

\textit{Temporal Grounding} questions ask when the event began and when it ended.

\textit{Location Grounding} questions focus on where the disaster is taking place, and the affected infrastructure.


Our multiple choice benchmarks are balanced, with roughly the same probability for each option to be the correct answer.






\textbf{Generated questions:} For the generated question-answer datasets, we prompt LLMs to create questions that are event specific, allowing for a more diverse variety of questions that pertain more specifically to the events in question.



\input{figures/qa_types}

\textbf{Train/test splits:}
We split the dataset by event to prevent location/temporal overlap. The train split contains 44,308 QA pairs, while the test set contains 10,196 QA pairs.

%% file: figures/geolocating_articles3.tex
\begin{figure}
    \centering
    \includegraphics[width=\linewidth]{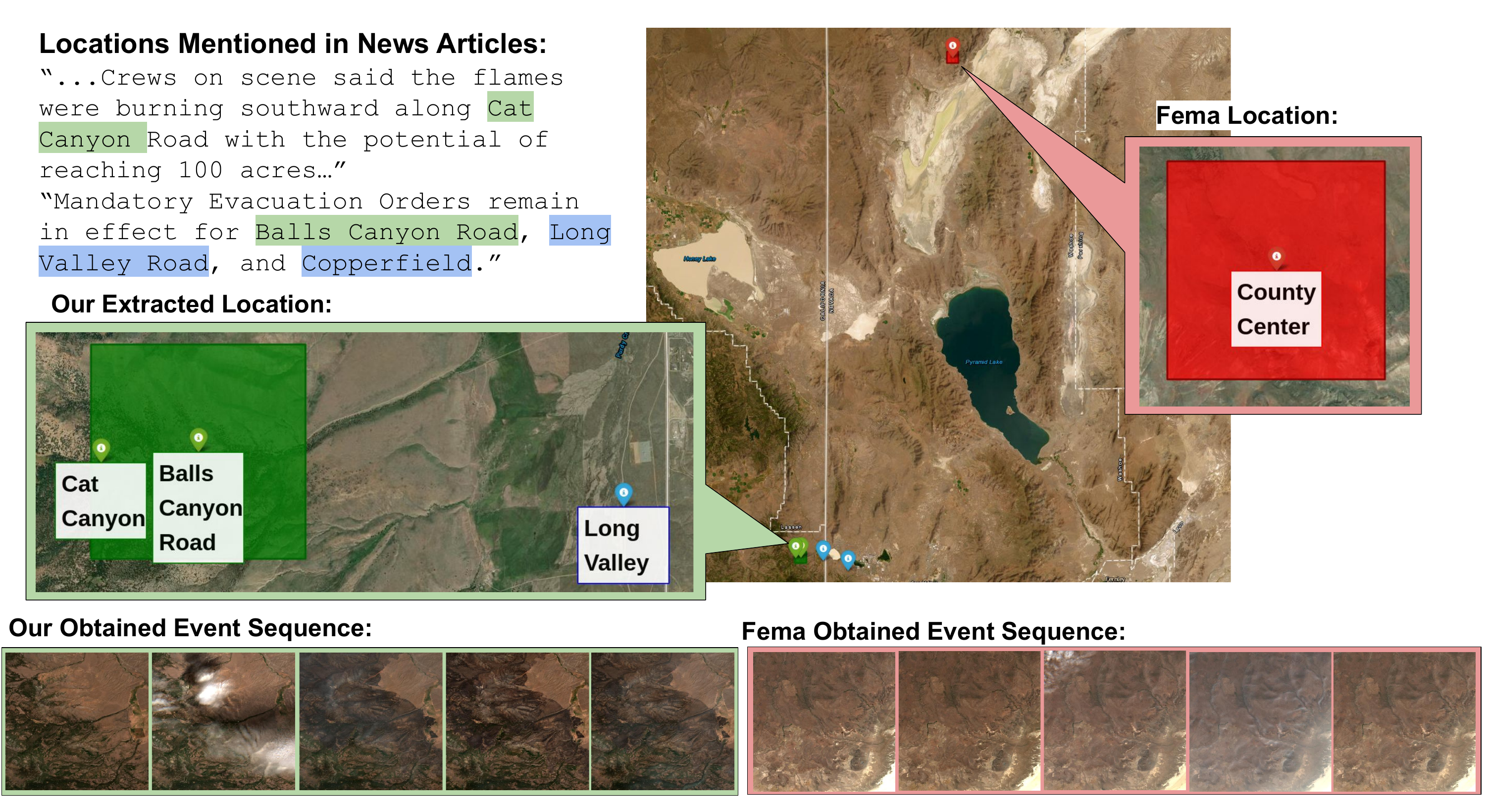}
    
   \caption{We demonstrate the use of geocoded news articles used to capture a better understanding of an events exact location. Here we visualize the result of our pipeline for the Loyalton Fire that took place in 2020, over the border of two neighboring states (California and Nevada). The FEMA provided coordinates for any event are the center of the county in which the event is located, however this does not necessarily provide the best coverage of the event, especially in cases like this where the disaster spans multiple counties, or in cases where the county is so large that the center coordinate is not near to the event location. Our sequence captures the progression of the fires by maintaining close distance to locations named in the news articles. }
    \label{fig:localization3}
\end{figure}

%% file: figures/caption_construction.tex
\begin{figure}[t]
    \centering
    \includegraphics[width=\linewidth]{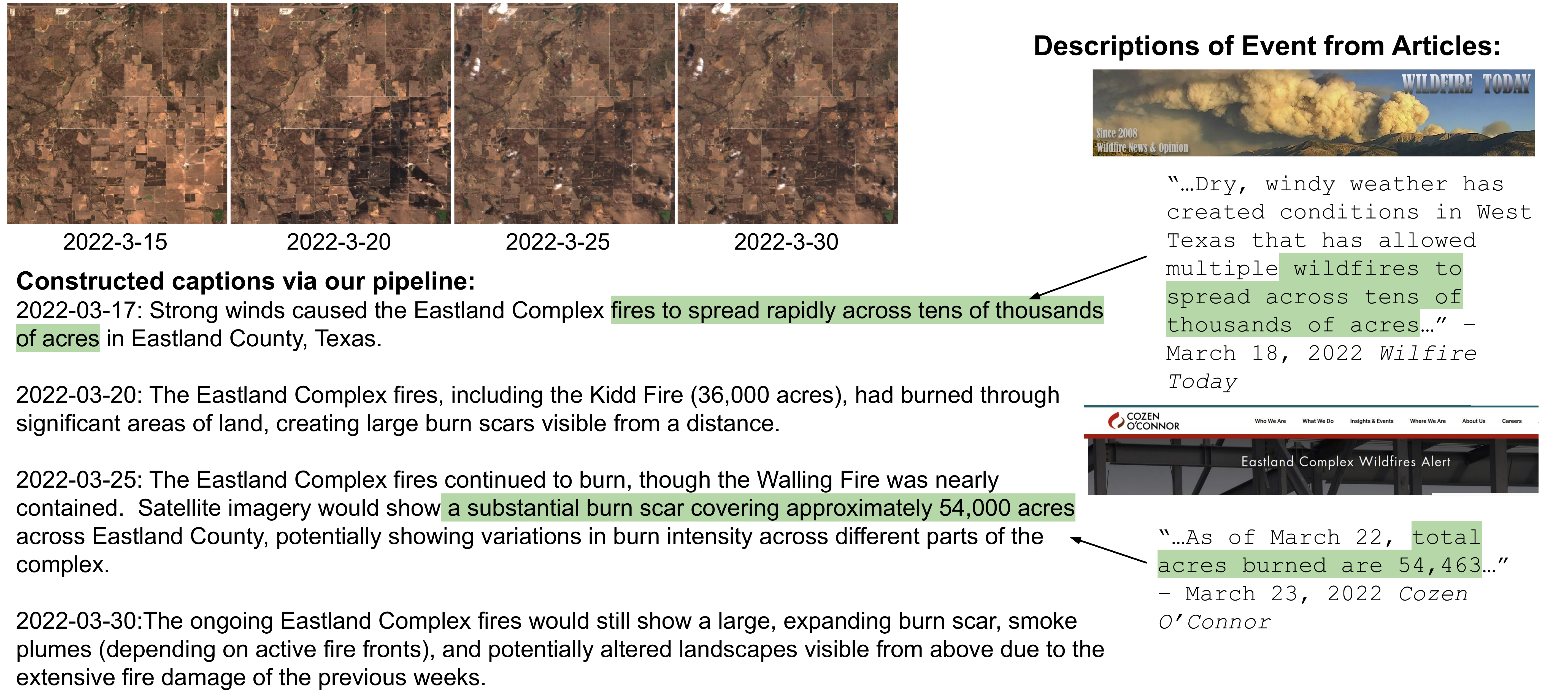}
    \caption{We illustrate the captions generated through our dataset construction pipeline. After geolocating the news articles, we prompt an LLM to retrieve captions using the articles' contents for a list of dates using the text alone. This ensures we are captioning the imagery independently of what may be visible. 
    We see that our process accurately describes the wildfire even in Eastland, Texas.}
    \label{fig:caption_example}
\end{figure}

%% file: figures/dataset_stats.tex
\begin{figure}[t]
    \centering
    \includegraphics[width=\linewidth]{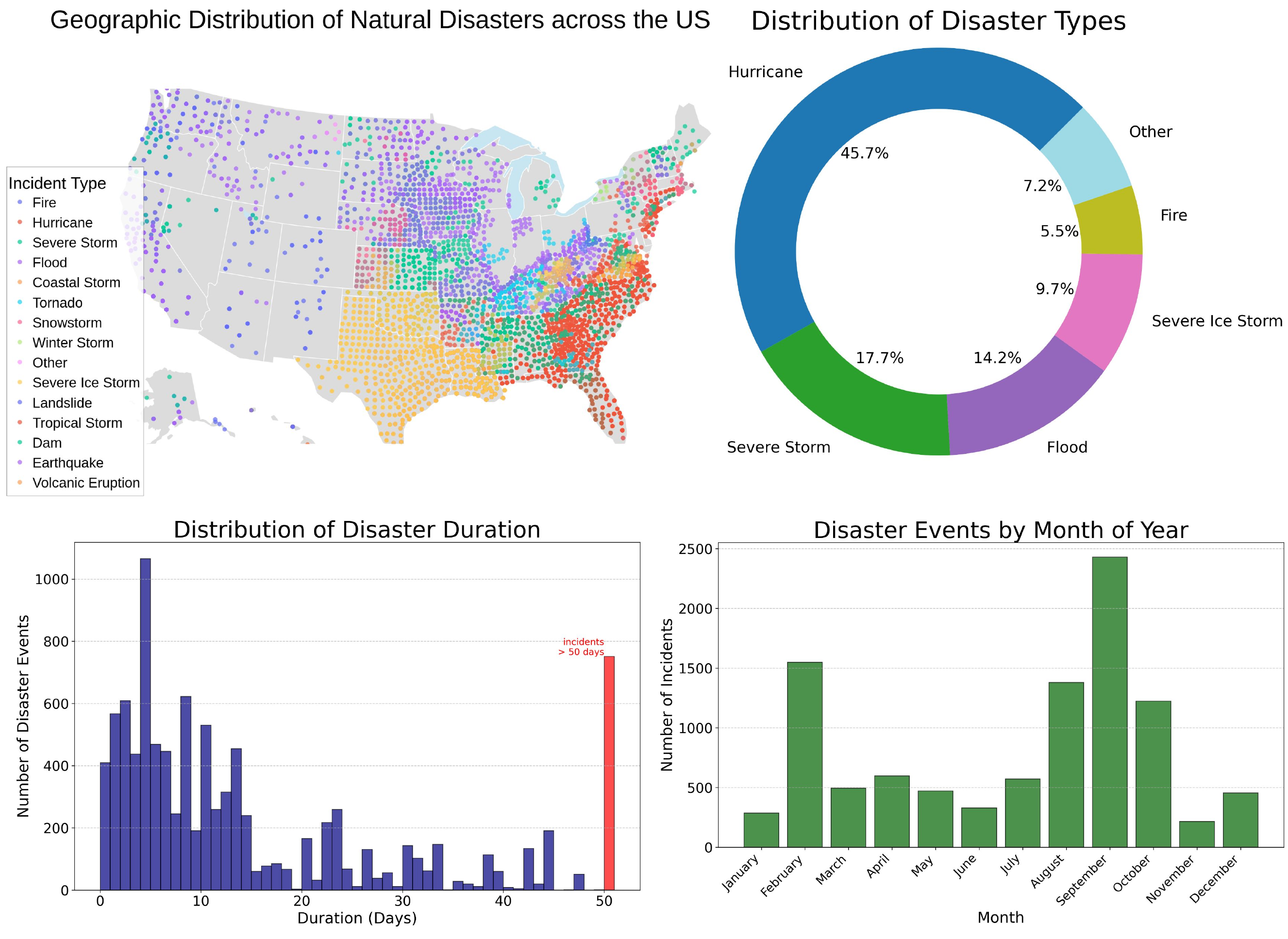}
    \caption{Our dataset represents the wide variety of natural disasters recorded by FEMA.}
    \label{fig:dataset_stats}
\end{figure}

%% file: figures/qa_types.tex
\begin{table}[t]
\label{tab:question_types}
\centering
\sloppy 
\footnotesize  
\setlength{\tabcolsep}{4pt}  
\renewcommand{\arraystretch}{1.15}
\begin{tabular}{p{1.5cm}p{1.8cm}p{3.0cm}p{6.5cm}}
\toprule
\textbf{Category} & \textbf{Question Type} & \textbf{Description} & \textbf{Example} \\
\midrule
\rowcolor[rgb]{0.9,1.0,0.9} 
Templated & Event Classification & Identifying which disaster is occurring & 
{\scriptsize What type of event is shown in these satellite images?
\begin{itemize}
\item[A:] [EVENT\_TYPE]
\item[B:] [EVENT\_TYPE]
\item[C:] [EVENT\_TYPE]
\item[D:] [EVENT\_TYPE]
\end{itemize}} \\
\rowcolor[rgb]{0.95,1.0,0.95}
Templated & Temporal Grounding & Determining when disasters begin and end & 
{\scriptsize Based on this sequence of satellite images from [DATES], which date shows the first evidence of the [EVENT\_TYPE]?} \\
\rowcolor[rgb]{0.9,1.0,0.9}
Templated & Location Grounding & Identifying where disasters occur and affected infrastructure & 
{\scriptsize What happened at [LOCATION] before [DATE]?} \\
\rowcolor[rgb]{0.85,0.9,1.0} 
Generated & Event-specific MCQ & Multiple choice questions with event-specific details & 
{\scriptsize Analyzing the progression of the wildfire, what appears to be the primary factor influencing its spread?
\begin{itemize}
\item[A:] Strong prevailing winds pushing the fire eastward.
\item[B:] The presence of a significant amount of dry brush and easily combustible vegetation.
\item[C:] Proximity to a major water source, significantly hindering fire spread.
\item[D:] Planned burns implemented by local fire departments effectively slowing the blaze.
\end{itemize}} \\
\rowcolor[rgb]{0.9,0.95,1.0}
Generated & Event-specific Free-response & Questions about specific events & 
{\scriptsize What were the conditions that led to the rapid spread of wildfires in Kansas, Texas, and Oklahoma?} \\
\bottomrule
\end{tabular}

\caption{Categorization of disaster-related questions in our dataset.}
\end{table}

%% file: sections/experiments.tex
\section{Experiments}


\paragraph{Experimental Setup}
For our baseline evaluation, we include the following models:
\begin{itemize}[noitemsep,topsep=0pt]
    \item VideoLLaVA 7b~\cite{lin2023video}: A video-language model that has been adapted for temporal reasoning tasks.
    \item TEOchat 7b~\cite{irvin2024teochat}: A recent multimodal model specifically designed for temporal earth observation data, which should theoretically be well-suited for our task.
    \item Gemini 2.0-flash~\cite{gemini}: A state-of-the-art closed-source multimodal model that has demonstrated strong performance on various vision-language tasks.
\end{itemize}
 We finetune TEOChat on our MONITRS-QA training set. We finetune for 1 epoch, with batch size of 4. Due to computational constraints, we conducted our experiments on a reduced training set (approximately 1/5th the size of our MONITRS-QA training dataset), with 1 epoch taking 3 hours trained on 3 A6000 GPUS.

\paragraph{Metrics}

For the multiple choice question-answer datasets we report overall accuracy and perform McNemar's statistical test~\cite{mcnemar1947note} to assess the significance of performance differences between models and validate observed improvements in MCQ tasks. For open-ended answers, we use established metrics for question-answering: BLEU~\cite{papineni2002bleu}, ROUGE-L~\cite{lin2004rouge}, and METEOR~\cite{banerjee2005meteor}, which measure n-gram overlap, longest common subsequence and semantic similarity respectively. Additionally we analyze answers using LLMs as judges, as described in Zheng et. al~\cite{zheng2023judging}. In general we ask Gemini 2.0-flash to score the factual accuracy, completeness, specificity, use of visual evidence, and the answer overall. We include the exact prompts in the appendix.


\section{Results}

We discuss quantitative results on MONITRS-QA in the main paper, while providing additional qualitative examples and visualizations of model predictions in the appendix.

\subsection{Multiple Choice Event Classification and Grounding}

\paragraph{Current state-of-the-art:}

Overall, we found baseline models struggle to answer questions related to natural disasters. For event classification, baseline performances hover around $\sim$50\%. Performance drops even lower for temporal~(11-18\%) and location~(13-17\%) grounding.

\paragraph{Results after finetuning on \datasetname-QA:}
Given the poor performance of current state-of-the-art, we finetune TEOchat~\cite{irvin2024teochat}, using a reduced training set (approximately 1/5th the size of our MONITRS-QA training dataset), with 1 epoch.

As shown in Table~\ref{tab:mcq_table}, our finetuned model significantly outperforms the baselines on all multiple-choice task types. For event classification, our model achieves 88.69\% accuracy, the gap widens further for temporal grounding, where our model achieves 70.72\% accuracy.

We conducted McNemar's test~\cite{mcnemar1947note} to assess the statistical significance of performance differences between models. Our finetuned model demonstrated statistically significant improvements over all baselines (p < 0.001). Specifically, our model correctly answered 296 questions that TEOChat missed for event classification (while TEOChat, the model specialized in temporal satellite events only correctly answered 11 questions our model missed).

\paragraph{Task-Specific Challenges:}We hypothesize that the gap between results in temporal grounding and event classification may be due to the idea that some events can be classified from a single image alone, but that temporal grounding which requires looking at the entire sequence, is not being learned.

Even with limited finetuning, the improvement for event classification and temporal grounding is both substantial and statistically significant (p < 0.01 to p < 0.001). This suggests that models are capable of learning to identify natural disasters, but have not quite learned to pick up on the gradual changes that are needed to differentiate types of events.

Location grounding remains challenging even for all models, but even then our finetuned model maintained statistically significant improvements over all baselines (p < 0.01 to p < 0.001). But it is important to note all models including ours struggled with this task, suggesting that additional sources of information such as location embeddings or segmentation masks are needed to properly locate concepts within imagery.

\begin{table}[t]
\centering
\caption{Multiple Choice Event Classification \& Grounding }
\scalebox{0.9}{
\begin{tabular}{l|c|c|c}
\hline
{Method} & Event Classification & Temporal Grounding & Location Grounding \\
\hline
Videollava~\cite{lin2023video} & 49.72\% &  11.11\% & 17.11\% \\
TEOchat~\cite{irvin2024teochat} & 48.88\% & 15.15\% & 15.50\% \\
Gemini 2.0-flash~\cite{gemini} & 50.07\% & 18.02\% & 13.74\% \\
Ours & 88.69\% & 70.72\%& 23.25\% \\
\hline
\end{tabular}
}   
\label{tab:mcq_table}
\end{table}

\subsection{General Disaster Response VQA}

From Table~\ref{tab:generated_table}, all models showed lower overall accuracy. Our fine-tuned model maintained significant advantages (52.18\% versus 28-37\% for baselines, p < 0.001), but the performance gap slightly narrowed compared to templated tasks. Our model correctly answered over 1000 questions that each baseline missed, while failing on only 362-431 questions where baselines succeeded. 
 
The results from the LLM-based evaluation in Table~\ref{tab:generated_table_eval}, suggest that fine-tuning on MONITRS improves the model's ability to connect language with visual features regarding natural disasters.

\begin{table}[t]
\centering
\caption{Generated VQA}
\scalebox{0.8}{
\begin{tabular}{l|c|ccccccc}
\hline
\multirow{2}{*}{Method} & {Multiple-Choice} & \multicolumn{6}{c}{Open-Ended} \\
 & Accuracy & BLEU-1 & BLEU-2 & BLEU-3 & BLEU-4 & METEOR & ROUGE-L \\
\hline
Videollava~\cite{lin2023video} & 36.65\% &  0.3447 & 0.2814 & 0.2490 & 0.2221 & 0.4739 & 0.3965\\
TEOchat~\cite{irvin2024teochat} & 36.99\% & 0.3439 & 0.2805 & 0.2483 & 0.2216 & 0.4736 & 0.3951\\
Gemini 2.0-flash~\cite{gemini} & 28.13\% & 0.2050 & 0.1398 & 0.1123 & 0.0920 & 0.3478 & 0.2419\\
Ours & 52.18\% &   0.4046  & 0.3351 &   0.2969 &  0.2667  & 0.4912&   0.4275 \\
\hline
\end{tabular}
}
\label{tab:generated_table}
\end{table}

\begin{table}[t]
\centering
\caption{Generated VQA -- LLM Evaluation}
\scalebox{0.75}{
\begin{tabular}{l|cccccc}
\hline
\multirow{2}{*}{Method} &  \multicolumn{6}{c}{Open-Ended} \\
 & Factual Accuracy & Completeness & Specificity & Visual Evidence & Uncertainty Handling & Overall\\
\hline
Videollava~\cite{lin2023video}& 3.41 & 3.46 &3.53 &2.27 & 4.26 & 3.08 \\
TEOchat~\cite{irvin2024teochat}& 3.39 & 3.45 & 3.52 & 2.28 & 4.31 & 3.08  \\
Gemini 2.0-flash~\cite{gemini}  & 2.44  & 2.10  & 2.04  & 2.00  & 4.15  & 2.13 \\
Ours &  3.84 &   3.54 &   3.72 &   2.50  &  4.29   & 3.08\\
\hline
\end{tabular}
}
\label{tab:generated_table_eval}
\end{table}

\section{Discussion}

Overall baseline performance on tasks regarding natural disasters is poor, indicating a gap our dataset addresses. The significant improvement after fine-tuning shows existing architectures can learn disaster recognition and temporal progression in satellite imagery when given specialized data. MONITRS provides this missing component by aligning language descriptions with visual evidence at specific temporal stages. 
While this dataset currently contains data regarding natural disasters, there is room for generalization as the geolocating of events is done using articles. Our methodology could potentially be extended to other domains with other events that are documented in news and lack sufficient visual annotations.

\paragraph{Future Applications.} The MONITRS dataset offers potential value beyond the immediate disaster classification and description tasks we've explored. Some promising directions include:
\begin{itemize}[noitemsep,topsep=0pt]
    \item Representation Learning: The aligned multimodal nature of MONITRS is well-suited for learning representations for change events, potentially creating embeddings that capture the semantic meaning of various disaster stages even without accompanying images.
    \item Architectural Innovations: Future work could explore new architectural components like date/time embeddings that explicitly encode temporal information in models, improving their ability to reason about disaster events through time.
\end{itemize}


\paragraph{Limitations.}
While we see a number of applications and models that could benefit from our dataset, there are still several limitations.
Our dataset relies on FEMA records, which only cover U.S. disasters. This limits generalization to global disaster events that may have different visual signatures. 
Our imagery is sourced from Sentinel-2~\cite{sentinel}, which has a 10m per pixel resolution and revisit period of approximately 5 days, which may miss critical stages in rapidly evolving disasters.

While we have taken steps to ensure annotation quality, the descriptions generated by LLMs based on news articles may not always accurately reflect what is visible in the satellite imagery. Finally, our dataset only includes RGB satellite imagery. Additional spectral bands or synthetic aperture radar (SAR) data could provide valuable information, especially for cloud-covered regions.

%% file: sections/conclusion.tex
\section{Conclusion}
We presented MONITRS, a novel multimodal dataset that pairs temporal satellite imagery of natural disasters with natural language descriptions derived from news articles. Our approach addresses a significant gap in existing disaster monitoring datasets by providing fine-grained temporal annotations and diverse disaster types.

%% file: sections/appendix.tex
\newpage
\appendix

\section{Qualitative Results}
\begin{figure}
    \centering
    \includegraphics[width=0.4\linewidth]{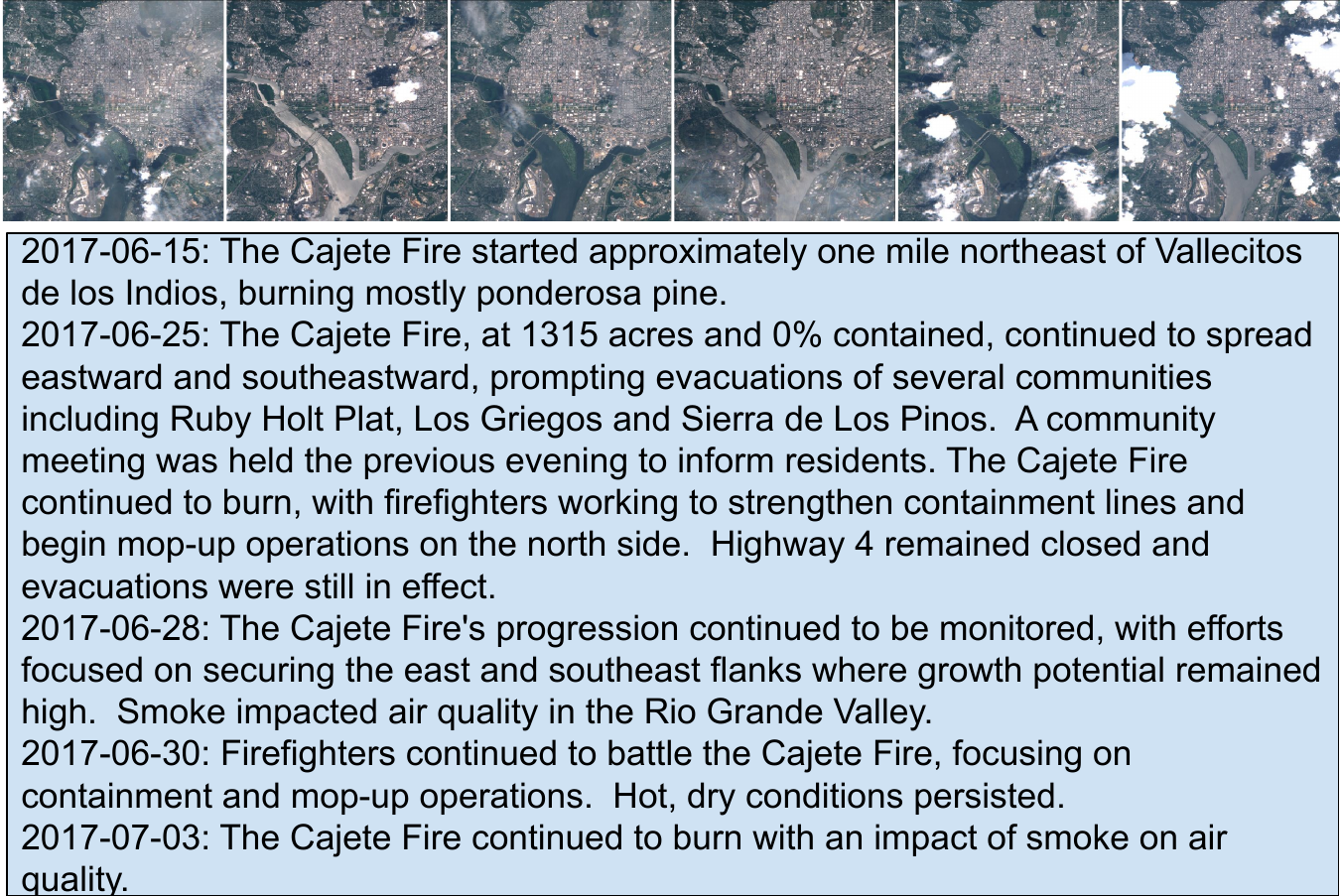}
    \includegraphics[width=0.4\linewidth]{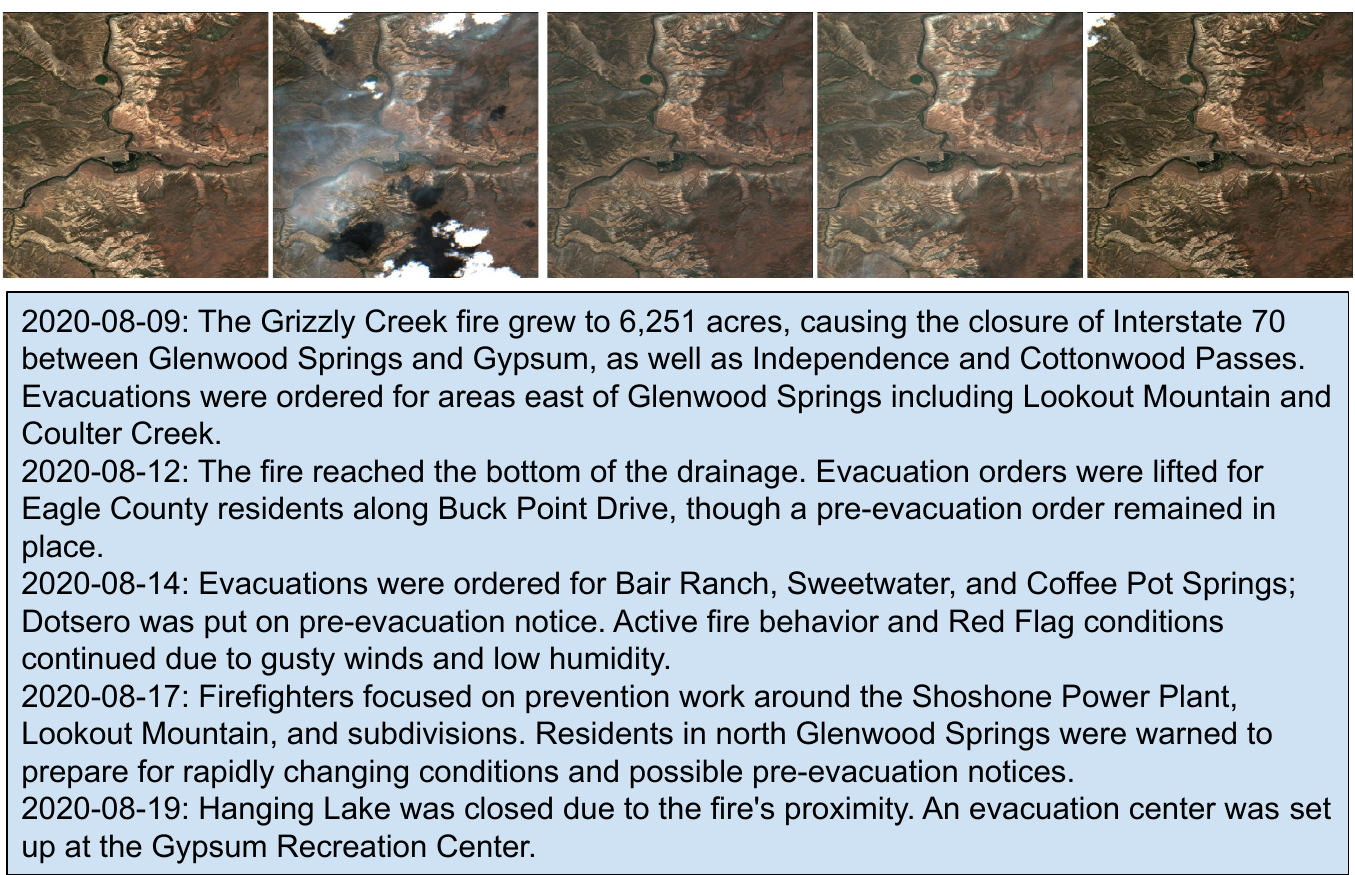}
    \includegraphics[width=0.4\linewidth]{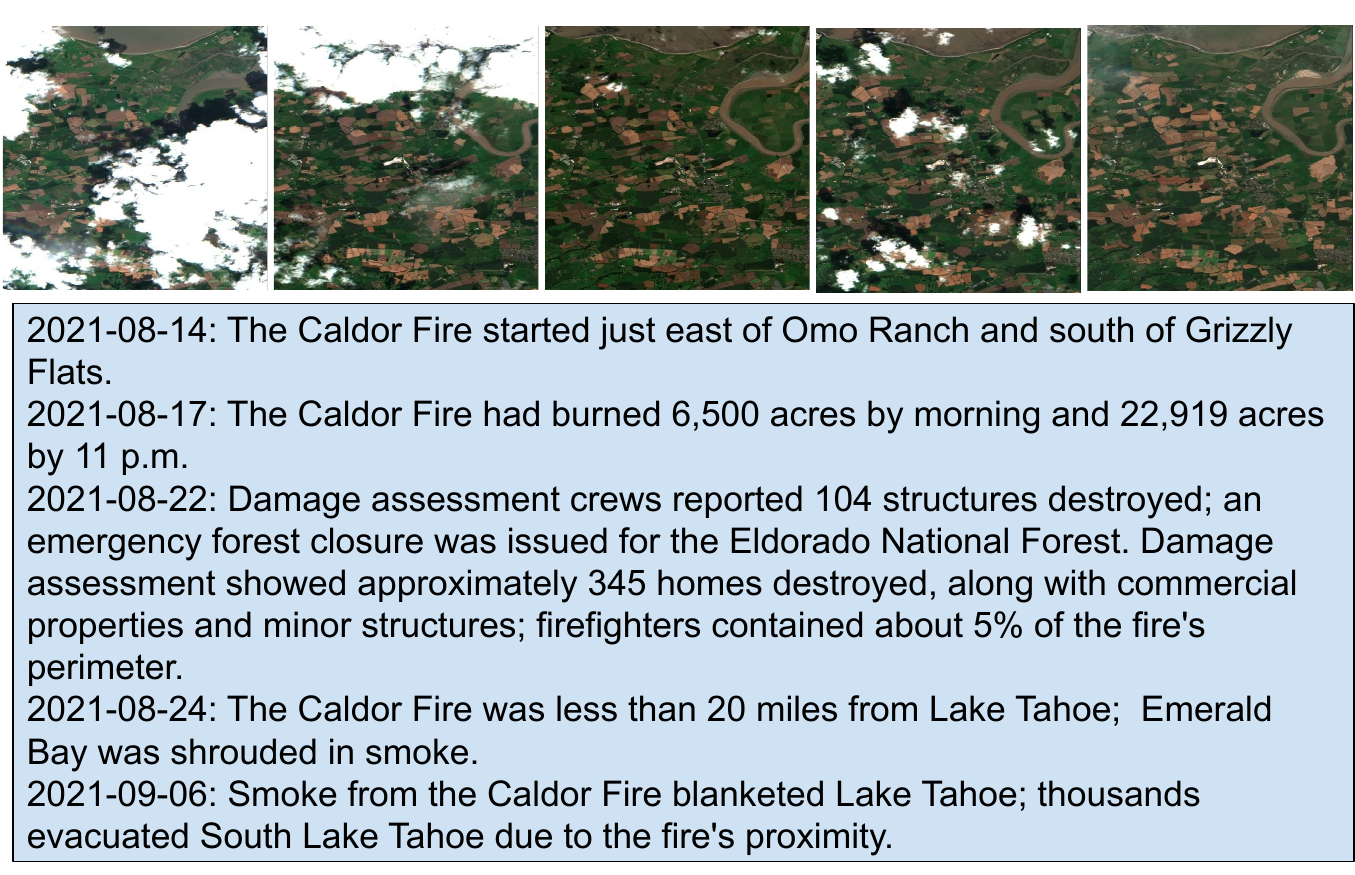}
    \includegraphics[width=0.4\linewidth]{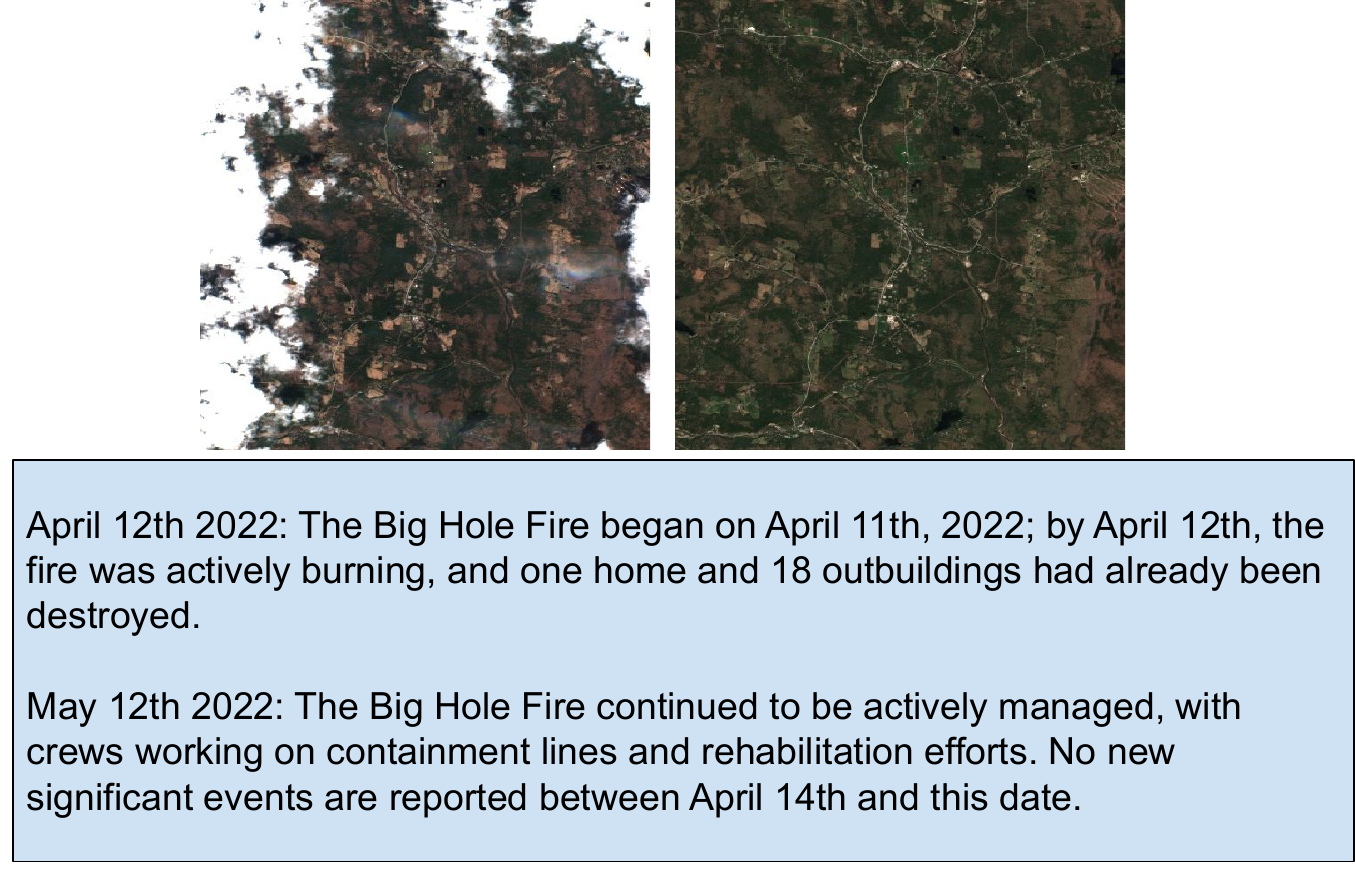}
    \includegraphics[width=0.4\linewidth]{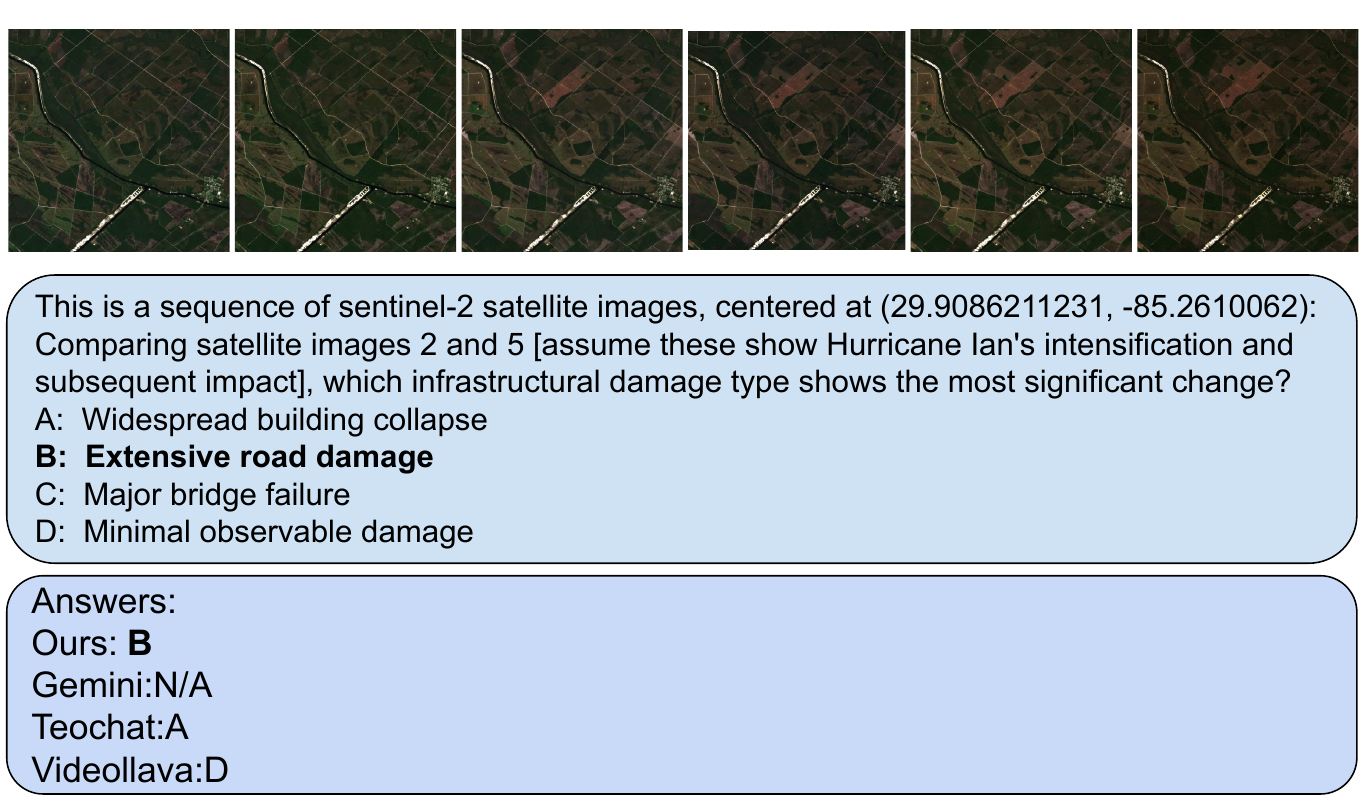}
    \includegraphics[width=0.4\linewidth]{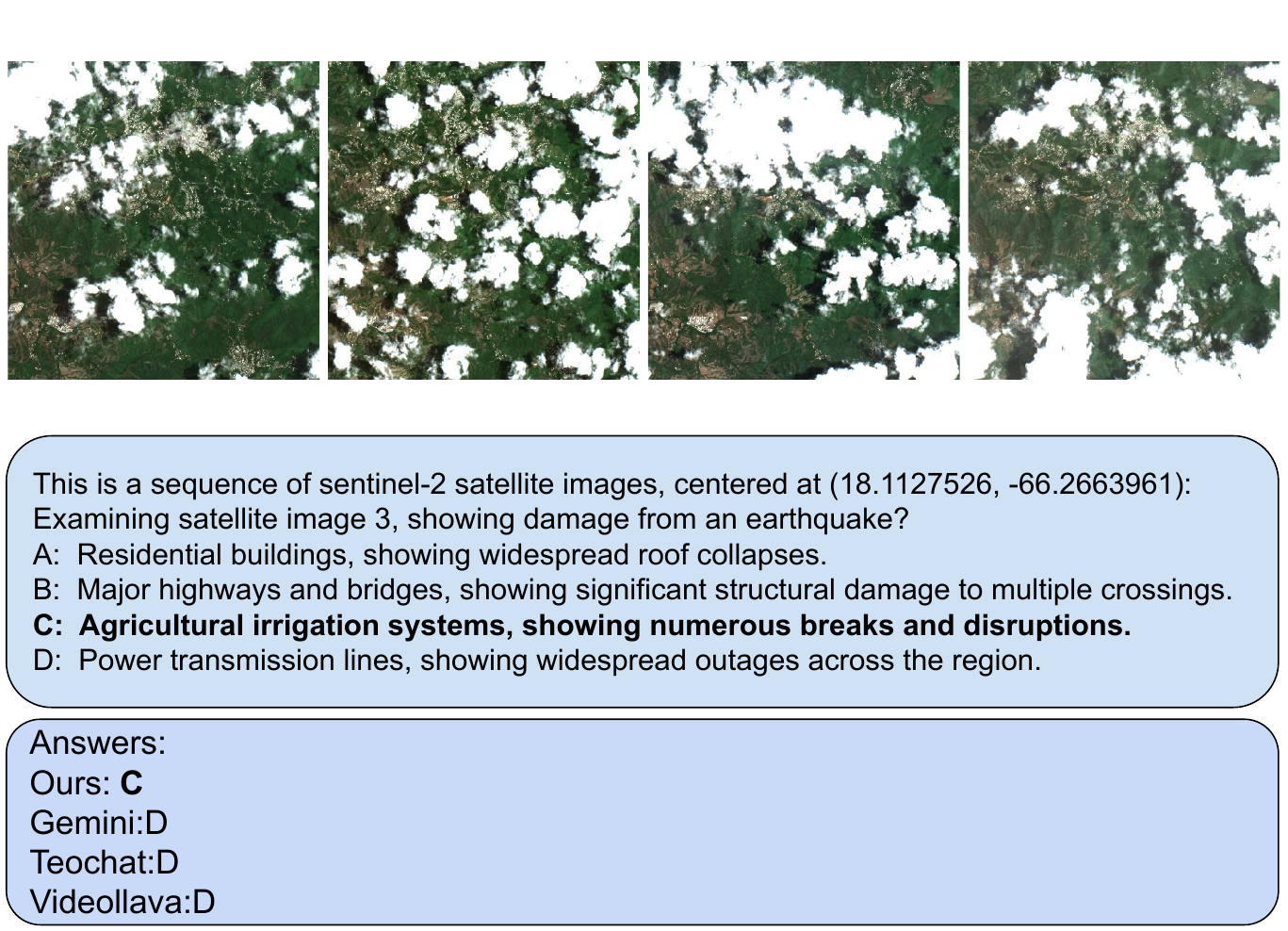}
    \includegraphics[width=0.4\linewidth]{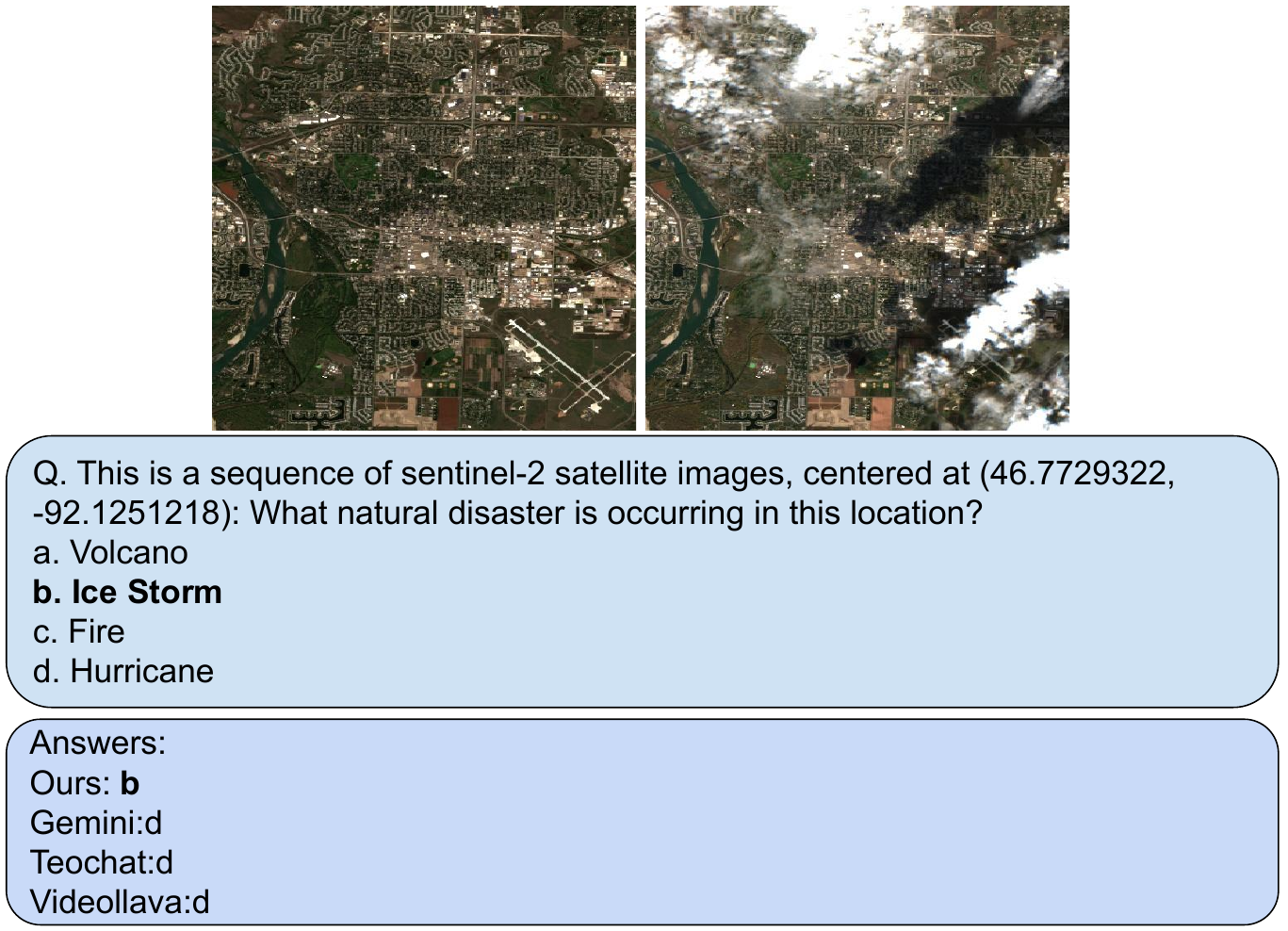}
    \includegraphics[width=0.4\linewidth]{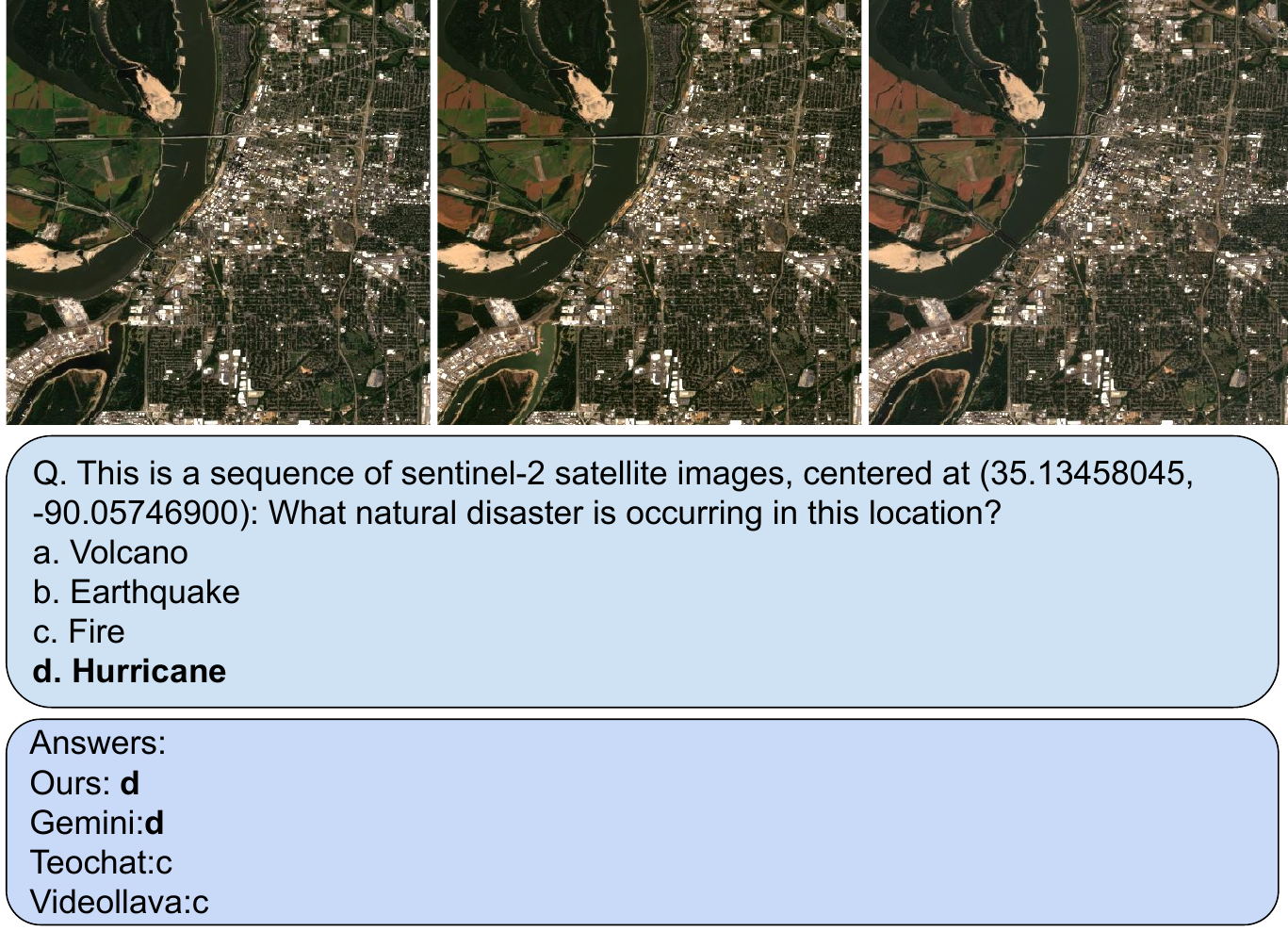}
    \includegraphics[width=0.4\linewidth]{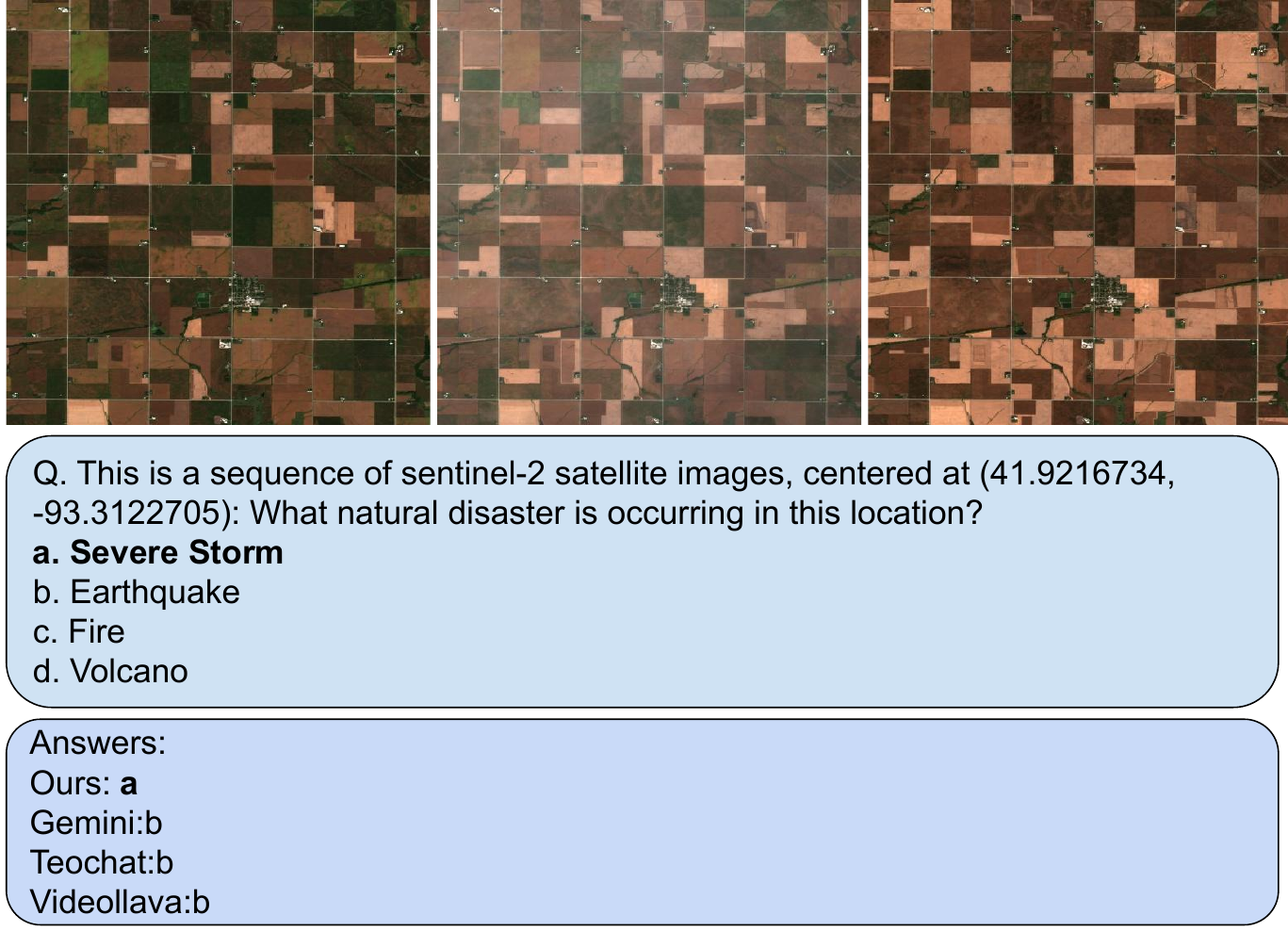}
    \includegraphics[width=0.4\linewidth]{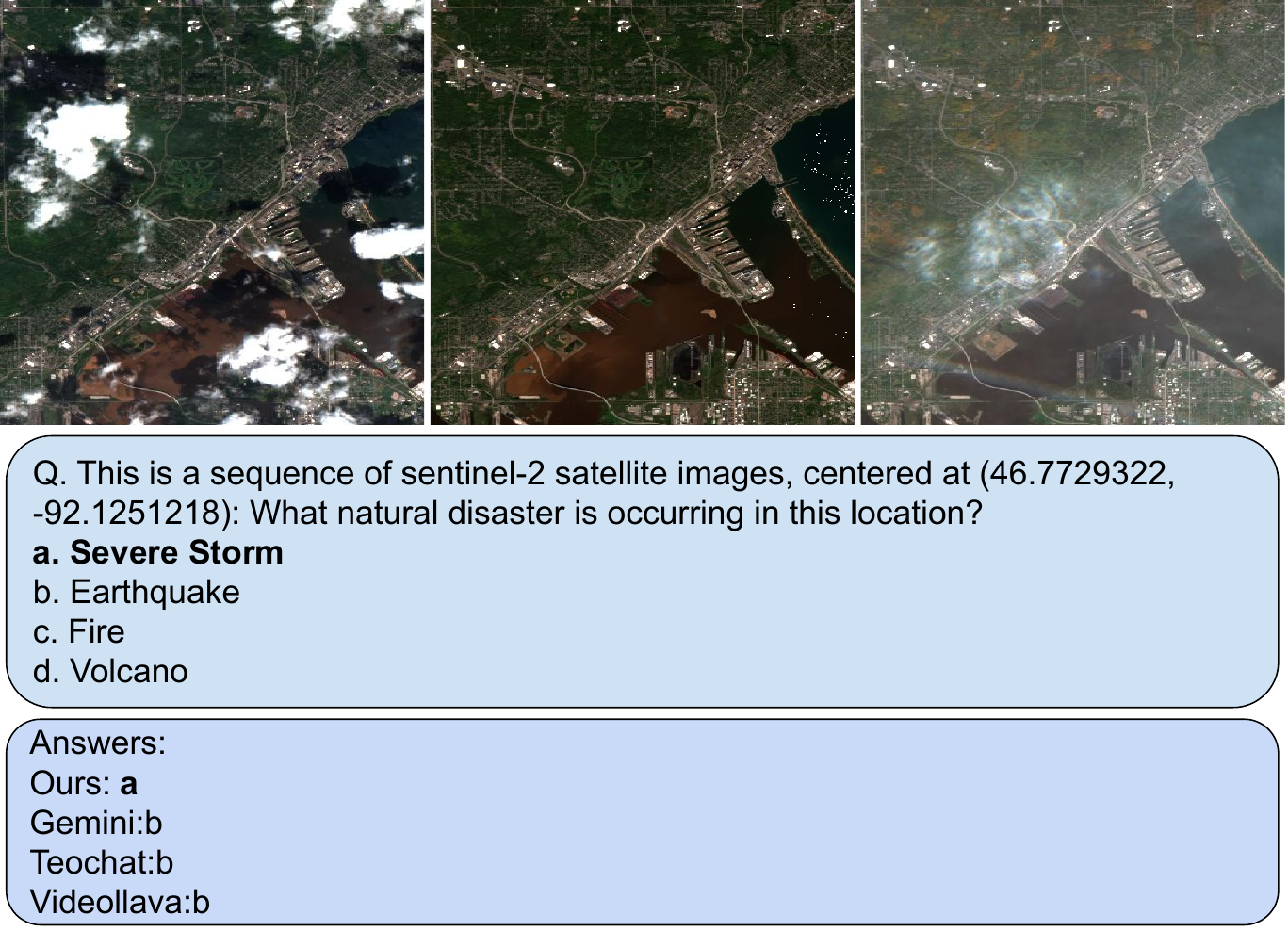}
    
    \caption{Qualitative examples from both MONITRS and MONITRS-QA along with their respective results.}
    \label{fig:qual}
\end{figure}

We include qualitative examples from both MONITRS and MONITRS-QA (along with results) in Figure~\ref{fig:qual}.

\section{Prompts to LLM}
We use prompts to LLMs to act as language tools for two types of tasks in our work.
The first being to read through and retrieve the relevant information from news articles to caption our image sequences, figures~\ref{fig:proper_noun_locations} and~\ref{fig:timeline-creation-prompt}
The second being utilizing our captions to generate event specific question-answer pairs, figures~\ref{fig:mcq-generation-prompt} and~\ref{fig:qa-generation-prompt}.

\begin{figure}
    \centering
    \begin{promptbox}
\textbf{Task:} Extract only the event-specific geographical locations mentioned in the provided articles about natural disasters.

\textbf{Instructions:}
\begin{enumerate}
    \item Carefully review the attached articles about natural disasters and identify ONLY proper noun locations that are directly related to where the disaster occurred or had direct impact.
    
    \item Focus on extracting:
    \begin{itemize}
        \item Specific sites where the event took place (cities, towns, neighborhoods)
        \item Precise natural features affected (specific rivers, mountains, forests, beaches)
        \item Particular infrastructure impacted (named dams, bridges, parks)
        \item Exact regions directly experiencing the disaster effects
    \end{itemize}
    
    \item Present your response in a simple string list format, with each location separated by a comma.
    
    \item If a location appears multiple times, include it only ONCE in your list.
    
    \item If the articles contain NO specific event locations, return only the word ``no'' (lowercase).
    
    \item DO NOT include:
    \begin{itemize}
        \item Broad geographical entities not directly affected (countries, states, unless the entire entity was impacted)
        \item Locations only mentioned incidentally (headquarters of responding agencies, etc.)
        \item Places mentioned for context but not directly experiencing the disaster
        \item General areas not specified with proper nouns
    \end{itemize}
\end{enumerate}

\textbf{Examples:}

For a wildfire article: \texttt{Paradise, Camp Creek Road, Butte County, Sierra Nevada foothills, Eastland County}

NOT: \texttt{California, United States, Western US}

For a hurricane article: \texttt{New Orleans, French Quarter, Lake Pontchartrain, Superdome}

NOT: \texttt{Louisiana, Gulf Coast, United States} (unless the entire state/region was directly impacted)

\textbf{Format for response when locations are found:}
\texttt{Paradise, Camp Creek Road, Butte County, Sierra Nevada foothills}

\textbf{Format for response when no locations are found:}
\texttt{no}

\textbf{Article Content:} \texttt{\{text\}}
\end{promptbox}
    \caption{Prompt given to LLM to extract proper nouns locations.}
    \label{fig:proper_noun_locations}
\end{figure}



\begin{figure}[htbp]
\centering
\begin{promptbox}
\textbf{Task:} Create a chronological timeline of observable natural disaster events from the provided news articles.

\textbf{Instructions:}
\begin{enumerate}
    \item Review the attached news articles for information about natural disasters (earthquakes, floods, hurricanes, wildfires, volcanic eruptions, etc.).
    
    \item For each date in the provided list, identify natural disaster events that occurred on or by that date that would be seen remotely.
    
    \item Write a 1-2 sentence description for each date focusing specifically on the visible physical manifestations, such as:
    \begin{itemize}
        \item Extent of flooding or inundation
        \item Wildfire burn scars or active fire fronts
        \item Hurricane cloud formations or aftermath flooding
        \item Visible structural damage to landscapes or urban areas
        \item Changes to coastlines, river courses, or terrain
        \item Ash clouds, lava flows, or other volcanic features
    \end{itemize}
    
    \item If a specific date isn't explicitly mentioned in the articles, use context clues to reasonably infer when these visible changes occurred.
    
    \item Present your response as a simple chronological list with dates followed by descriptions.
    
    \item Emphasize the VISUAL aspects that would be detectable from above.
\end{enumerate}

\textbf{Format example:}

\texttt{June 15, 2023: Extensive flooding covered approximately 60 square miles of the Mississippi Delta region, with standing water clearly visible across previously inhabited areas and farmland.}

\texttt{July 3, 2023: The Caldor wildfire in California created a distinct burn scar spanning 25 miles along the Sierra Nevada mountain range, with active fire fronts visible on the northeastern perimeter.}

\textbf{Article Content:} \texttt{\{text\}}

\textbf{Dates for analysis:} \texttt{\{dates\}}
\end{promptbox}
\caption{Prompt for creating chronological timelines of visually observable natural disaster events}
\label{fig:timeline-creation-prompt}
\end{figure}



\begin{figure}[htbp]
\centering
\begin{promptbox}
Given a set of statements in an order I'd like you to make 3 multiple choice questions about the events described.
Make the questions diverse, covering different aspects of the events that could be answerable using satellite imagery of the event.
Each question should have 4 options (A, B, C, and D) with only one correct answer.

\textbf{Statements:} \texttt{\textbackslash n\{events\}}

Format your response exactly like this:

\texttt{**Question 1:** [Your first question here]}
\texttt{A) [First option]}
\texttt{B) [Second option]}
\texttt{C) [Third option]}
\texttt{D) [Fourth option]}
\texttt{**Correct Answer 1:** [Correct option letter]}

\texttt{**Question 2:** [Your second question here]}
\texttt{A) [First option]}
\texttt{B) [Second option]}
\texttt{C) [Third option]}
\texttt{D) [Fourth option]}
\texttt{**Correct Answer 2:** [Correct option letter]}

\texttt{**Question 3:** [Your third question here]}
\texttt{A) [First option]}
\texttt{B) [Second option]}
\texttt{C) [Third option]}
\texttt{D) [Fourth option]}
\texttt{**Correct Answer 3:** [Correct option letter]}

\textbf{Here are some examples of statements:} 2021-12-11: No events described in the article are visible from this date. 2021-12-15: Very strong winds in Kansas, Texas, and Oklahoma caused numerous wildfires to spread rapidly. Blowing dust severely reduced visibility, causing streetlights to turn on at midday in some areas. 2021-12-16: A large wildfire in Russell and Ellis Counties, Kansas burned approximately 365,850 acres, destroying at least 10 homes. High winds, gusting up to 100 mph, fueled the fire and other blazes across western Kansas, Oklahoma, and Texas. 2021-12-21: No events described in the article are visible from this date.

\textbf{Here are some examples of questions:}

\texttt{**Question 1:** What natural disaster is visible in the satellite images from mid-December 2021?}
\texttt{A) Hurricane}
\texttt{B) Tornado}
\texttt{C) Wildfire}
\texttt{D) Flooding}
\texttt{**Correct Answer 1:** C}

\texttt{**Question 2:** Approximately how many acres were burned in Russell and Ellis Counties, Kansas?}
\texttt{A) 36,585 acres}
\texttt{B) 365,850 acres}
\texttt{C) 3,658 acres}
\texttt{D) 3,658,500 acres}
\texttt{**Correct Answer 2:** B}

\texttt{**Question 3:** What weather condition contributed significantly to the spread of wildfires in December 2021?}
\texttt{A) Heavy rainfall}
\texttt{B) Strong winds}
\texttt{C) Freezing temperatures}
\texttt{D) High humidity}
\texttt{**Correct Answer 3:** B}
\end{promptbox}
\caption{Prompt for generating multiple choice questions from natural disaster event statements}
\label{fig:mcq-generation-prompt}
\end{figure}

\begin{figure}[htbp]
\centering
\begin{promptbox}
Given a set of statements in an order I'd like you to make 3 questions about the events described.
Make the questions diverse, covering different aspects of the events that could be aided answerable using satellite imagery of the event.

\textbf{Statements:} \texttt{\textbackslash n\{events\}}

Format your response exactly like this:

\texttt{**Question 1:** [Your first question here]}
\texttt{**Answer 1:** [Your first answer as a complete sentence]}
\texttt{**Question 2:** [Your second question here]}
\texttt{**Answer 2:** [Your second answer as a complete sentence]}
\texttt{**Question 3:** [Your third question here]}
\texttt{**Answer 3:** [Your third answer as a complete sentence]}

\textbf{Here are some examples of statements:} 2021-12-11: No events described in the article are visible from this date. 2021-12-15: Very strong winds in Kansas, Texas, and Oklahoma caused numerous wildfires to spread rapidly. Blowing dust severely reduced visibility, causing streetlights to turn on at midday in some areas. 2021-12-16: A large wildfire in Russell and Ellis Counties, Kansas burned approximately 365,850 acres, destroying at least 10 homes. High winds, gusting up to 100 mph, fueled the fire and other blazes across western Kansas, Oklahoma, and Texas. 2021-12-21: No events described in the article are visible from this date. 2021-12-26: No events described in the article are visible from this date. 2021-12-31: No events described in the article are visible from this date. 2022-01-05: No events described in the article are visible from this date. 2022-01-10: No events described in the article are visible from this date. 2022-01-15: No events described in the article are visible from this date.

\textbf{Here are some examples of questions:}

\texttt{**Question 1:** What were the conditions that led to the rapid spread of wildfires in Kansas, Texas, and Oklahoma?}
\texttt{**Answer 1:** The conditions that led to the rapid spread of wildfires in Kansas, Texas, and Oklahoma were very strong winds, low humidity, and high temperatures.}

\texttt{**Question 2:** What was the impact of the wildfires in Russell and Ellis Counties, Kansas?}
\texttt{**Answer 2:** The impact of the wildfires in Russell and Ellis Counties, Kansas was the burning of approximately 365,850 acres and the destruction of at least 10 homes.}

\texttt{**Question 3:** When did the wildfires in Kansas, Texas, and Oklahoma occur?}
\texttt{**Answer 3:** The wildfires in Kansas, Texas, and Oklahoma occurred on December 15, 2021.}
\end{promptbox}
\caption{Prompt for generating question-answer pairs from natural disaster event statements}
\label{fig:qa-generation-prompt}
\end{figure}

%% file: neurips_2025.bbl
\begin{thebibliography}{10}

\bibitem{banerjee2005meteor}
Satanjeev Banerjee and Alon Lavie.
\newblock Meteor: An automatic metric for mt evaluation with improved correlation with human judgments.
\newblock In {\em Proceedings of the acl workshop on intrinsic and extrinsic evaluation measures for machine translation and/or summarization}, pages 65--72, 2005.

\bibitem{braik2024automated}
Abdullah~M Braik and Maria Koliou.
\newblock Automated building damage assessment and large-scale mapping by integrating satellite imagery, gis, and deep learning.
\newblock {\em Computer-Aided Civil and Infrastructure Engineering}, 39(15):2389--2404, 2024.

\bibitem{chen2024application}
Zhonghan Chen.
\newblock Application of uav remote sensing in natural disaster monitoring and early warning: an example of flood and mudslide and earthquake disasters.
\newblock {\em Highlights in Science, Engineering and Technology}, 85:924--933, 2024.

\bibitem{dong2024changeclip}
Sijun Dong, Libo Wang, Bo~Du, and Xiaoliang Meng.
\newblock Changeclip: Remote sensing change detection with multimodal vision-language representation learning.
\newblock {\em ISPRS Journal of Photogrammetry and Remote Sensing}, 208:53--69, 2024.

\bibitem{sentinel}
ESA.
\newblock Copernicus sentinel-2, 2024.

\bibitem{fema_dis}
FEMA.
\newblock National response framework, 2025.

\bibitem{fema_set}
Federal Emergency Management~Agency (FEMA).
\newblock Openfema dataset, 2024.

\bibitem{galetto2024use}
Federico Galetto, Diego Lobos~Lillo, and Matthew~E Pritchard.
\newblock The use of high-resolution satellite topographic data to quantify volcanic activity at raung volcano (indonesia) from 2000 to 2021.
\newblock {\em Bulletin of Volcanology}, 87(1):1--19, 2025.

\bibitem{gemini}
Google.
\newblock Gemini.

\bibitem{GoogleCustomSearchAPI}
{Google Developers}.
\newblock Custom search json api.
\newblock \url{https://developers.google.com/custom-search/v1/overview}, 2025.
\newblock Last updated: 2025-05-07, Accessed: 2025-05-16.

\bibitem{gupta2019xbd}
Ritwik Gupta, Richard Hosfelt, Sandra Sajeev, Nirav Patel, Bryce Goodman, Jigar Doshi, Eric Heim, Howie Choset, and Matthew Gaston.
\newblock xbd: A dataset for assessing building damage from satellite imagery.
\newblock {\em arXiv preprint arXiv:1911.09296}, 2019.

\bibitem{hoxha2022change}
Genc Hoxha, Seloua Chouaf, Farid Melgani, and Youcef Smara.
\newblock Change captioning: A new paradigm for multitemporal remote sensing image analysis.
\newblock {\em IEEE Transactions on Geoscience and Remote Sensing}, 60:1--14, 2022.

\bibitem{irvin2024teochat}
Jeremy~Andrew Irvin, Emily~Ruoyu Liu, Joyce~Chuyi Chen, Ines Dormoy, Jinyoung Kim, Samar Khanna, Zhuo Zheng, and Stefano Ermon.
\newblock Teochat: A large vision-language assistant for temporal earth observation data.
\newblock {\em arXiv preprint arXiv:2410.06234}, 2024.

\bibitem{kuckreja2024geochat}
Kartik Kuckreja, Muhammad~Sohail Danish, Muzammal Naseer, Abhijit Das, Salman Khan, and Fahad~Shahbaz Khan.
\newblock Geochat: Grounded large vision-language model for remote sensing.
\newblock In {\em Proceedings of the IEEE/CVF Conference on Computer Vision and Pattern Recognition}, pages 27831--27840, 2024.

\bibitem{li2024vision}
Xiang Li, Congcong Wen, Yuan Hu, Zhenghang Yuan, and Xiao~Xiang Zhu.
\newblock Vision-language models in remote sensing: Current progress and future trends.
\newblock {\em IEEE Geoscience and Remote Sensing Magazine}, 2024.

\bibitem{lin2023video}
Bin Lin, Yang Ye, Bin Zhu, Jiaxi Cui, Munan Ning, Peng Jin, and Li~Yuan.
\newblock Video-llava: Learning united visual representation by alignment before projection.
\newblock {\em arXiv preprint arXiv:2311.10122}, 2023.

\bibitem{lin2004rouge}
Chin-Yew Lin.
\newblock Rouge: A package for automatic evaluation of summaries.
\newblock In {\em Text summarization branches out}, pages 74--81, 2004.

\bibitem{liu2022remote}
Chenyang Liu, Rui Zhao, Hao Chen, Zhengxia Zou, and Zhenwei Shi.
\newblock Remote sensing image change captioning with dual-branch transformers: A new method and a large scale dataset.
\newblock {\em IEEE Transactions on Geoscience and Remote Sensing}, 60:1--20, 2022.

\bibitem{liu2024remoteclip}
Fan Liu, Delong Chen, Zhangqingyun Guan, Xiaocong Zhou, Jiale Zhu, Qiaolin Ye, Liyong Fu, and Jun Zhou.
\newblock Remoteclip: A vision language foundation model for remote sensing.
\newblock {\em IEEE Transactions on Geoscience and Remote Sensing}, 2024.

\bibitem{mall-23}
Utkarsh Mall, Bharath Hariharan, and Kavita Bala.
\newblock Change-aware sampling and contrastive learning for satellite images.
\newblock In {\em CVPR}, 2023.

\bibitem{mall2023remote}
Utkarsh Mall, Cheng~Perng Phoo, Meilin~Kelsey Liu, Carl Vondrick, Bharath Hariharan, and Kavita Bala.
\newblock Remote sensing vision-language foundation models without annotations via ground remote alignment.
\newblock {\em ICLR}, 2024.

\bibitem{manas2021seasonal}
Oscar Manas, Alexandre Lacoste, Xavier Gir{\'o}-i Nieto, David Vazquez, and Pau Rodriguez.
\newblock Seasonal contrast: Unsupervised pre-training from uncurated remote sensing data.
\newblock In {\em Proceedings of the IEEE/CVF International Conference on Computer Vision}, pages 9414--9423, 2021.

\bibitem{manas-21}
Oscar Manas, Alexandre Lacoste, Xavier Gir{\'o}-i Nieto, David Vazquez, and Pau Rodriguez.
\newblock Seasonal contrast: Unsupervised pre-training from uncurated remote sensing data.
\newblock In {\em ICCV}, 2021.

\bibitem{geocodemapsco}
{Map Maker}.
\newblock Geocoding api.
\newblock \url{https://geocode.maps.co/}, 2025.
\newblock Accessed: 2025-05-16.

\bibitem{mcnemar1947note}
Quinn McNemar.
\newblock Note on the sampling error of the difference between correlated proportions or percentages.
\newblock {\em Psychometrika}, 12(2):153--157, 1947.

\bibitem{noman2024remote}
Mubashir Noman, Mustansar Fiaz, Hisham Cholakkal, Sanath Narayan, Rao~Muhammad Anwer, Salman Khan, and Fahad~Shahbaz Khan.
\newblock Remote sensing change detection with transformers trained from scratch.
\newblock {\em IEEE Transactions on Geoscience and Remote Sensing}, 2024.

\bibitem{papineni2002bleu}
Kishore Papineni, Salim Roukos, Todd Ward, and Wei-Jing Zhu.
\newblock Bleu: a method for automatic evaluation of machine translation.
\newblock In {\em Proceedings of the 40th annual meeting of the Association for Computational Linguistics}, pages 311--318, 2002.

\bibitem{park2024national}
Jongsoo Park, Hagyu Jeong, and Junwoo Lee.
\newblock National disaster management and monitoring using satellite remote sensing and geo-information.
\newblock {\em Korean Journal of Remote Sensing}, 40(5):813--832, 2024.

\bibitem{revankar2024scale}
Shreelekha Revankar, Cheng~Perng Phoo, Utkarsh Mall, Bharath Hariharan, and Kavita Bala.
\newblock Scale-aware recognition in satellite images under resource constraints.
\newblock {\em arXiv preprint arXiv:2411.00210}, 2024.

\bibitem{sachdeva2023change}
Ragav Sachdeva and Andrew Zisserman.
\newblock The change you want to see.
\newblock In {\em Proceedings of the IEEE/CVF Winter Conference on Applications of Computer Vision}, pages 3993--4002, 2023.

\bibitem{said2019natural}
Naina Said, Kashif Ahmad, Michael Riegler, Konstantin Pogorelov, Laiq Hassan, Nasir Ahmad, and Nicola Conci.
\newblock Natural disasters detection in social media and satellite imagery: a survey.
\newblock {\em Multimedia Tools and Applications}, 78:31267--31302, 2019.

\bibitem{sheykhmousa2019post}
Mohammadreza Sheykhmousa, Norman Kerle, Monika Kuffer, and Saman Ghaffarian.
\newblock Post-disaster recovery assessment with machine learning-derived land cover and land use information.
\newblock {\em Remote sensing}, 11(10):1174, 2019.

\bibitem{tanim2022flood}
Ahad~Hasan Tanim, Callum~Blake McRae, Hassan Tavakol-Davani, and Erfan Goharian.
\newblock Flood detection in urban areas using satellite imagery and machine learning.
\newblock {\em Water}, 14(7):1140, 2022.

\bibitem{thangavel2023autonomous}
Kathiravan Thangavel, Dario Spiller, Roberto Sabatini, Stefania Amici, Sarathchandrakumar~Thottuchirayil Sasidharan, Haytham Fayek, and Pier Marzocca.
\newblock Autonomous satellite wildfire detection using hyperspectral imagery and neural networks: A case study on australian wildfire.
\newblock {\em Remote Sensing}, 15(3):720, 2023.

\bibitem{yang2024made}
Charig Yang, Weidi Xie, and Andrew Zisserman.
\newblock Made to order: Discovering monotonic temporal changes via self-supervised video ordering.
\newblock In {\em European Conference on Computer Vision}, pages 268--286. Springer, 2024.

\bibitem{zhang2023video}
Hang Zhang, Xin Li, and Lidong Bing.
\newblock Video-llama: An instruction-tuned audio-visual language model for video understanding.
\newblock {\em arXiv preprint arXiv:2306.02858}, 2023.

\bibitem{zheng2023judging}
Lianmin Zheng, Wei-Lin Chiang, Ying Sheng, Siyuan Zhuang, Zhanghao Wu, Yonghao Zhuang, Zi~Lin, Zhuohan Li, Dacheng Li, Eric Xing, et~al.
\newblock Judging llm-as-a-judge with mt-bench and chatbot arena.
\newblock {\em Advances in Neural Information Processing Systems}, 36:46595--46623, 2023.

\bibitem{zhu2024semantic}
Yongshuo Zhu, Lu~Li, Keyan Chen, Chenyang Liu, Fugen Zhou, and Zhenwei Shi.
\newblock Semantic-cc: Boosting remote sensing image change captioning via foundational knowledge and semantic guidance.
\newblock {\em IEEE Transactions on Geoscience and Remote Sensing}, 2024.

\end{thebibliography}
